\documentclass[10pt, a4paper]{article}

\usepackage{lrec-coling2024} 

\usepackage{todonotes}
\usepackage{dblfloatfix}
	
\usepackage{multirow}

\usepackage{tabularx}
\usepackage[utf8]{inputenc}
\usepackage[russian,english]{babel}
\usepackage{tablefootnote}

\usepackage{arydshln}

\title{Multilingual Substitution-based Word Sense Induction}

\name{Denis Kokosinskii\textsuperscript{1}, Nikolay Arefyev\textsuperscript{2}} 

\address{\textsuperscript{1}Lomonosov Moscow State University, Russia \\
        \textsuperscript{2}SaluteDevices, Russia \\
         \textsuperscript{3}University of Oslo, Norway   \\
         deniskokosss@mail.ru, nikolare@uio.no \\
         }

\abstract{
Word Sense Induction (WSI) is the task of discovering senses of an ambiguous word by grouping usages of this word into clusters corresponding to these senses. Many approaches were proposed to solve WSI in English and a few other languages, but these approaches are not easily adaptable to new languages. We present multilingual substitution-based WSI methods that support any of 100 languages covered by the underlying multilingual language model with minimal to no adaptation required. 
Despite the multilingual capabilities, our methods perform on par with the existing monolingual approaches on popular English WSI datasets. At the same time, they will be most useful for lower-resourced languages which miss lexical resources available for English, thus, have higher demand for unsupervised methods like WSI .
 \\ \newline  }

\begin{document}

\maketitleabstract

\section{Introduction}
The task of word sense induction (WSI) is to cluster occurrences (\textit{instances}) of an ambiguous word into clusters corresponding to the senses of this word. 
E.g., for 
(1) \textit{He sat on the {\bf bank} of the river}, (2) \textit{He cashed a check at the {\bf bank}}, and (3) \textit{That {\bf bank} holds the mortgage on my home}, the desired clusters are \{1\} and \{2, 3\}. 

The core assumption in WSI is the lack of a predefined sense inventory. If a word sense inventory is available, the clustering task can be reformulated as a classification task over senses in this inventory. It is usually referred to as Word Sense Disambiguation (WSD) in this case. It is worth noticing that WSI is more relevant than WSD for lower-resourced languages where linguistic resources are scarce.

Although many approaches exist for WSI, the SOTA or near-SOTA ones are language-specific and are not easily adaptable to new languages. Our main goal is to create a WSI system that works seamlessly in many languages. The contributions of this paper are the following.

1. Multilingual WSI methods are proposed that generalize to any language supported by the underlying multilingual language model with minimal\footnote{For 78 languages supported by both the Stanza \citep{qi2020stanza} lemmatizer and the XLM-R multilingual LM~\citep{XLMR} only one Hearst-like pattern should be translated.} to no adaptation while performing on par with the state-of-the-art WSI approaches for English\footnote{Code is available at \url{https://github.com/deniskokosss/mwsi.git} .}.

2. We perform rigorous evaluation and analysis of various configurations of our system on datasets in 11 languages.

3. We discover remarkable abilities of a multilingual masked language model after monolingual unsupervised finetuning to generate cross-lingual lexical substitutes performing on par with the monolingual ones for WSI.

\section{Related Work}
WSI can be formulated as either a hard clustering problem, meaning each instance must be put to a single cluster, or a soft clustering problem where each instance must receive some weights or probabilities of belonging to each cluster. SemEval 2010 task 14 \citeplanguageresource{manandhar-etal-2010-semeval} and SemEval 2013 task 13 \citeplanguageresource{jurgens-klapaftis-2013-semeval} are the standard WSI benchmarks in English requiring hard and soft clustering, respectively. RUSSE'2018 \citeplanguageresource{Russe}, and RuDSI \citeplanguageresource{rudsi} are publicly available WSI benchmarks in Russian requiring hard clustering.

To the best of our knowledge, no prior work has attempted to design a multilingual WSI model. However, recently in~\citelanguageresource{XL-WSD} a benchmark and several methods for multilingual WSD have been introduced. Their multilingual methods rely on XLM-R~\citep{XLMR}, a Transformer-based masked language model trained on a corpus of 100 languages. For WSD methods labeled training data for each word sense and/or sense definitions are required. This limits the applicability of the proposed methods for lower-resourced languages.

\begin{figure*}[h]
\includegraphics[width=\linewidth]{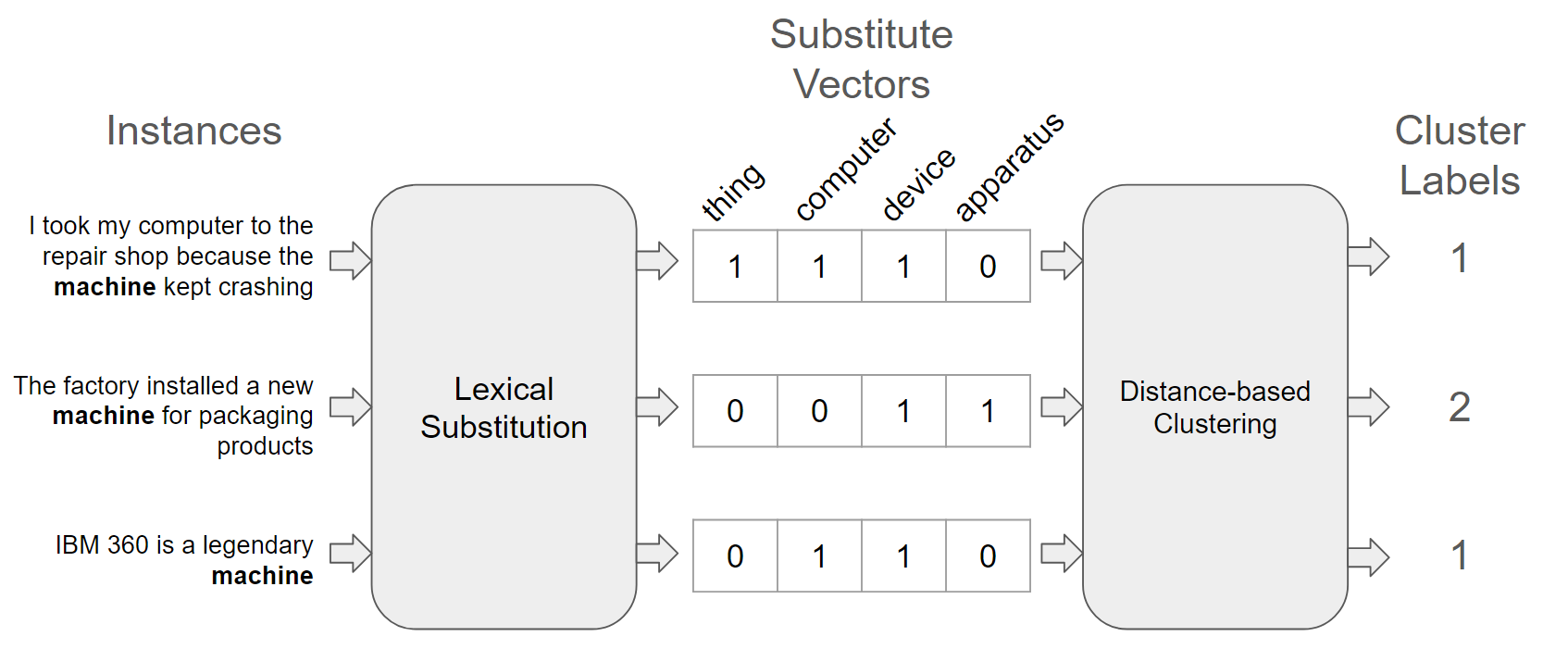}
\caption{Substitution-based approach to Word Sense Induction. }
\label{fig:lexsub_framework}
\end{figure*}

As for monolingual WSI methods, one popular approach implemented in~\citep{lau2013unimelb, wang2015sense, komninos2016structured, amplayo2019autosense} relies on the Latent Dirichlet Allocation method of topic modeling. AutoSense \citep{amplayo2019autosense}, the latest and the best-performing method in this group, incorporates topic and sense latent variables, as well as a switch variable to determine whether to use the local context of the target word or the whole instance.

The performance of WSI methods improved significantly with the adoption of neural masked language models. Recently, PolyLM \citep{ansell2021polylm} introduced a transformer-based architecture for learning sense embeddings from an unlabeled corpus. PolyLM allocates eight embeddings for each word to represent up to eight different senses per word. The model is trained to make these embeddings distinct and helpful for the standard masked language modeling task. An important detail is that PolyLM relies on a word-level vocabulary. However, a multilingual word-level vocabulary will be prohibitively large and it is not clear how this model can be adapted to work with a subword-level vocabulary. 

Another class of methods for WSI, including AI-KU \citep{bacskaya2013ai}, LSDP ELMo \citep{amrami2018word}, LSDP BERT \citep{amrami2019towards}, BERT NoDP (\citealp{amrami2019towards}; \citealp{eyal2022large}) and +embs \citep{arefyev2020always}, employ lexical substitution. A \textit{lexical substitute} is a word that can replace some target word in a given sentence without making the sentence ungrammatical, and also have the same or similar/related meaning. These methods use a two-step process (see Figure \ref{fig:lexsub_framework}): first, generate a vector of lexical substitutes for each instance using a substitution model and then cluster the resulting substitute vectors. An important advantage of lexical substitution methods for WSI is their interpretability. Each cluster can be labeled with a few substitutes that describe this particular sense. It is worth noticing that lexical substitution is a long-standing NLP task by itself. There are several human-annotated benchmarks for lexical substitution: COInCo \citeplanguageresource{CoInCo} and SemEval2007 task 10 \citeplanguageresource{Se07} in English, GermEval 2015 \citeplanguageresource{GermEval} in German, and SemDisFr 2014 \citeplanguageresource{SemDis} in French. We employ these datasets for intrinsic comparison of our lexical substitution methods and studying the relations between target words and lexical substitutes generated by these methods. Next we describe the process of generating substitutes in two of the substitution-based WSI methods, BERT LSDP \citep{amrami2019towards} and +embs \citep{arefyev2020always}. These methods show SOTA results on English WSI benchmarks.

BERT LSDP (Language model Substitutions with Dynamic Patterns) follows the substitution-based approach to WSI and uses BERT to produce lexical substitutes. In theory, BERT can estimate probabilities of lexical substitutes by simply replacing the target word with the <mask> token. However, \citet{amrami2019towards} show that such substitutes are too general and often unrelated to the target word. BERT LSDP overcomes this issue and "injects" the information about the target word into substitutes. Specifically, it averages two vectors of unnormalized probabilities (logits) of potential substitutes. The first vector of logits is taken from the BERT MLM head directly at the first token of the target word in a given instance without masking it. The second vector of logits is taken from the <mask> token after replacing the target word with the \textit{dynamic pattern} "<T> (or even <mask>)". For example, for the sentence "This cat is cute" the input to BERT after applying the dynamic pattern is "This cat (or even <mask>) is cute." \citealp{xlmr-russe} have shown that combining substitutes from two \textit{symmetric dynamic patterns}, e.g. "<T> (or even <mask>)" and "<mask> (or even <T>)" consistently improves WSI performance compared to using one of these patterns alone. This was demonstrated for several patterns on a WSI dataset in Russian.

 The substitution-based method "+embs" \citep{arefyev2020always} relies on non-contextualized embeddings to inject the target word. The "general" substitutes are first produced by a language model with the target word masked. These substitutes are then reranked based on the apriori similarity to the target word. More specifically, the probabilities of substitutes to appear in the given context as estimated by a masked language model are multiplied by the exponentiated and re-normalized cosine similarities between their non-contextualized embeddings and the non-contextualized embedding of the target word. Several underlying language models were tested with +embs. XLNET \citep{yang2019xlnet} has shown superior performance, while the results with BERT are comparable to BERT LDSP. 

\section{Substitution-based Methods for Multilingual WSI}


\subsection{From BERT to XLM-R}\label{31}

Incorporating a multilingual language model into BERT-based methods for WSI seems straightforward. XLM-R \citep{XLMR} is a publicly available Masked Language Model (MLM) trained on a corpus of 100 languages. Theoretically, it can replace BERT \citep{BERT} in the monolingual methods and make them multilingual. However, we found the results after this replacement unsatisfactorily low and had to analyze the underlying reasons to improve the quality.

\begin{figure}
\centering
\includegraphics[width=\linewidth]{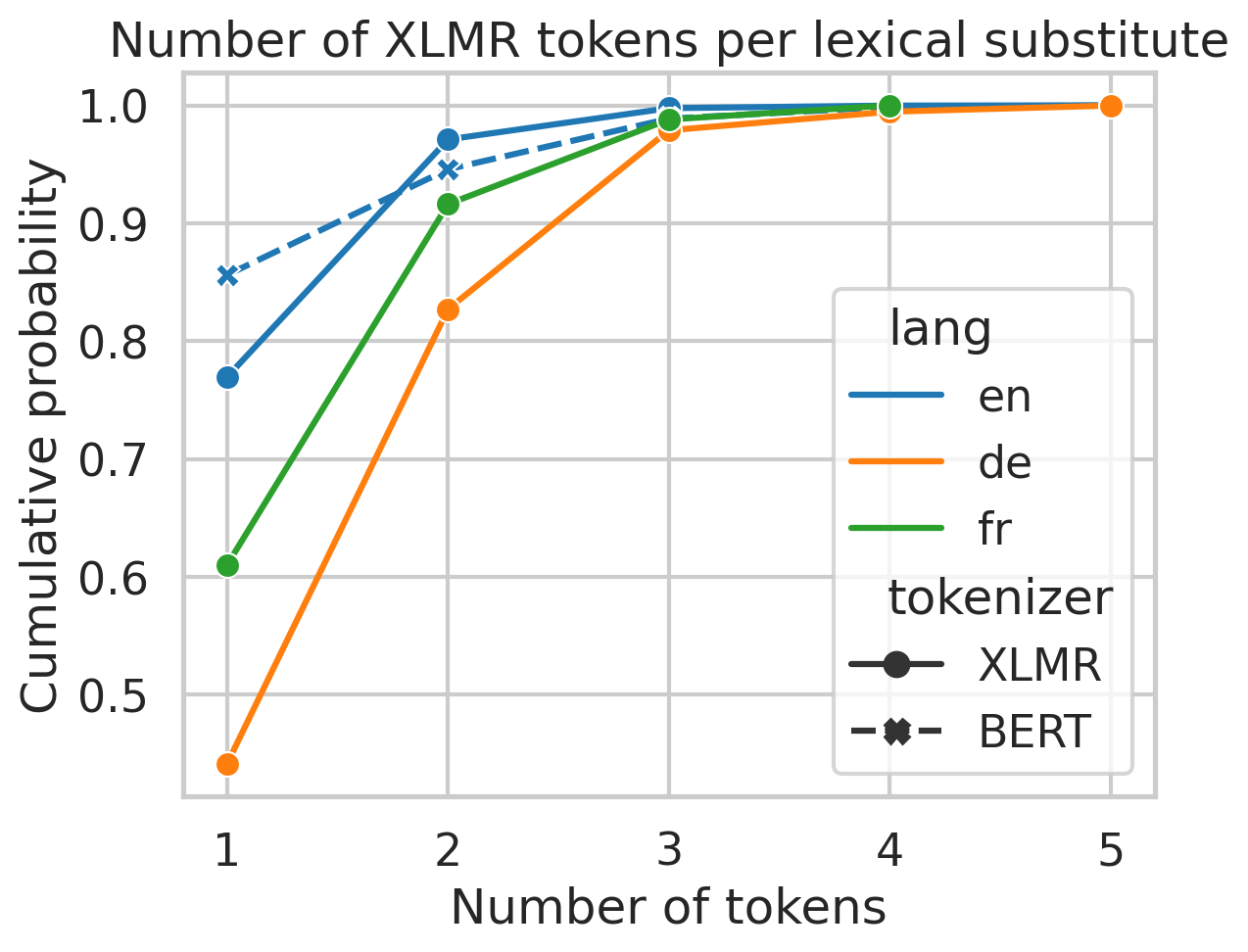}
\caption{Cumulative distribution of the number of XLM-R and BERT tokens per substitute in human-annotated lexical substitution datasets. English substitutes are from CoInCo \citep{CoInCo}, German substitutes are from GermEval 2015 \citep{GermEval}, and French substitutes are from SemDis 2014 \citep{SemDis}. Multi-word substitutes are not taken into account.}
\label{fig:tokens_per_subst}
\end{figure}

The main reason is the difference in tokenization between BERT and XLM-R. Despite having a significantly larger token vocabulary (250k in XLM-R versus 30k in English BERT), the average number of tokens per word is significantly higher on average for XLM-R due to its multilinguality. Figure \ref{fig:tokens_per_subst} shows the distribution of the number of XLM-R tokens per word for human-written lexical substitutes in English, French, and German substitution datasets. While only approximately 15\% of English substitutes consist of multiple tokens in BERT, for XLM-R this number increases to 20\% and reaches 40\% for French substitutes and 55\% for German substitutes. The original BERT LSDP method only produces single-subword substitutes, thus limiting the diversity of substitutes yielded by XLM-R. \citet{xlmr-russe} have shown that two-subword XLM-R substitutes achieve significantly better WSI results for Russian. In general we want all reasonable substitutes to be generated independently of their length. Thus, we introduce methods that produce substitutes of variable length. 

Employing the multilingual and cross-lingual capabilities of XLM-R, we experiment with generation of substitutes in the language of the input word occurrence, and also cross-lingual substitutes in English or Russian independently of the input language.


\subsubsection{Concat} 
In the \textit{Concat} substitute generator substitutes are generated by a masked language model that receives an instance with a target word replaced by 1, 2, or 3 consecutive <mask> tokens.\footnote{We do not generate substitutes consisting of more than 3 subwords because such substitutes are very rarely suggested by humans as figure~\ref{fig:tokens_per_subst} shows. Experimental results on the importance of the used number of masks for WSI are presented in Appendix \ref{app:num_masks}. }
For example, an instance "\textbf{cats} are cute" turns into three input sequences:

"<mask> are cute", 

"<mask><mask> are cute",

"<mask><mask><mask> are cute". 

For each input we select $k$ most probable tokens for the first mask and decode tokens for the following masks using greedy decoding\footnote{We tried using beam search instead, but did not observe significant improvements.}, resulting in $k$ generated sequences of subwords. We then post-process these sequences by merging subwords into words and keeping only the first word from each sequence\footnote{We do not use the whole multi-word expressions generated because they increase the sparsity of the substitute vectors without providing intuitive benefits.}. The resulting probability of a substitute is the product of conditional probabilities of its subwords. In total, $3k$ substitutes are generated and $k$ most probable are selected among them. 

\subsubsection{WCM} 

The Word Continuation Masking (WCM) approach requires self-supervised finetuning of the original masked language model, unlike concat. The idea of WCM is to train the masked language model to produce tokens of a lexical substitute one at a time, similarly to autoregressive language models, while leveraging both left and right contexts during generation. This kind of finetuning can be done on an unlabeled corpus of texts.

The autoregressive generation of substitutes cannot be adequately implemented for masked language models out of the box. For instance, in case when only one mask is present (e.g., "cats and <mask> are cute") only single-token substitutes will likely be generated by an MLM. The first tokens of multi-token words will have low probabilities because the model "sees" only one mask and "understands" that there is no space for a multi-token word. 

WCM is similar to the standard masked language modeling objective with an important difference. Similarly to BERT 15\% of tokens in an input text are masked and the objective is to restore the masked tokens. However, in WCM the boundaries of the words fully or partially replaced by a <mask> are then identified. All tokens of such words following the <mask> token are removed. The tokens that precede the <mask> token and the tokens of other words are left intact. Thus, the model is trained to predict the first token of a masked word or to continue a word without knowing the exact length of this word. During inference, masks get inserted into the context one by one to generate substitutes of arbitrary length. For example, the following decoding path is possible: "cats and <mask> are cute", "cats and capy<mask> are cute", "cats and capybara<mask> are cute", resulting in a three-token substitute "capybaras".

We trained \textbf{WCM}, \textbf{WCM-en}, and \textbf{WCM-ru}, three models based on XLM-R and finetuned with the WCM objective. WCM is trained on texts in all 100 languages supported by XLM-R, preserving the original proportions of languages, while WCM-en is trained only on texts in English and WCM-ru only on texts in Russian\footnote{We trained each model for 2 epochs on $\approx$6GB of texts randomly sampled from CommonCrawl100 \citeplanguageresource{CC100}.}.

\subsubsection{Monolingual fine-tuning results in cross-lingual substitutes} 

We empirically observe a remarkable property of monolingual WCM finetuning. WCM-en and WCM-ru always yield substitutes in the finetuning language, not in the language of the input text. Surprisingly, these substitutes often constitute a reasonable description of the meaning of the target word. For example, for a sentence \textit{à la <mask> comme à la guerre!}\footnote{A French proverb \textit{à la guerre comme à la guerre!}, translates as \textit{at war as at war}.} the top-5 substitutes by WCM-en are words in English that are semantically similar to the masked target word in French: \textit{war, peace, liberty, faith, freedom}. The same behavior holds for more complex examples, such as the ones found in WSI datasets.

The possibility of generating sensible English substitutes for a word occurrence in any language allows using English lemmatizer instead of searching for a lemmatizer for each input language. This significantly simplifies adaptation of our WSI methods to new languages.
Cross-lingual substitutes in English may also enable direct utilization of rich English knowledge resources for other languages, but we leave this for future research. It also provides interpretable sense labels in English for an ambiguous word in any language. We provide examples of such labels in section 5.

\subsection{Target Injection}\label{32}
A straightforward approach to generate substitutes with a masked language model is to replace the target word with a mask. This way, a language model produces substitutes that fit the context of a sentence. However, these substitutes may not be semantically related to the original target word, because the target word itself is removed from the sentence. Target injection aims to condition the language model substitutes with the target word. We adapt two target injection methods, described in Section 2, to multilingual applications.

\textbf{SDP (Symmetric Dynamic Patterns).} Due to the intuitive nature of dynamic patterns for target injection, they are straightforward to adapt for multilingual applications. We select the "<T> (or even <mask>)" pattern with its symmetric counterpart "<mask> (or even <T>)" for our experiments following \citep{amrami2019towards}. We have experimented with several other patterns and observed no significant differences in WSI performance (refer to Appendix \ref{app:hyper_sel} for details). We use Google Translate to translate the patterns to the target language. It is the only action required to adapt our methods to any of the 100 XLM-R pretraining languages. We also simplify the transition between languages even further and experiment with the non-translated versions of the dynamic patterns for WCM-en and WCM-ru, relying on their cross-lingual capabilities.

In our implementation of the target injection with SDP, the probabilities of substitutes for two symmetric patterns are multiplied with smoothing\footnote{If a substitute is not present in $k$ substitutes of one of the patterns, it is assigned a probability of 1e-5.}. Thus, the substitutes in the top $k$ for both dynamic patterns will rise to the top of the ranking. 

\textbf{+embs.} In the original +embs method, the substitutes can only consist of a single BERT token. It is limiting in the case of XLM-R, as discussed previously. More than that, the original method uses the embedding layer of the MLM as the non-contextualized embeddings of tokens. This approach is also not straightforward to adapt for the case of multi-token substitutes.

We use FastText \citep{fasttext} non-contextualized embeddings instead of relying on the embedding layer of the MLM. Fasttext provides out-of-the-box non-contextualized embeddings on all the languages from XLM-R pretraining. We compute the distances between the FastText embeddings of the target word and each of the $3k$ most probable language-model substitutes yielded by a multi-token substitute generator. As in the original method, all the $3k$ non-contextualized similarities pass through a temperature softmax. It helps align the "peaky" distribution from the substitute generator with a more "flat" distribution of FastText similarities. The final ranking of substitutes is the product of the non-contextualized (FastText) and the contextualized (MLM) distributions.

\subsection{Multilingual WSI Methods}

Our methods for multilingual WSI follow the substitution-based approach. We generate $k$ substitutes of 1 to 3 XLM-R tokens with one of two multi-token substitute generators, described in Section \ref{31}. We use one of the methods from Section \ref{32} for target injection. We see no theoretical preference in any of the given options and rely on experiments to determine the best configuration. 

Intuitively, a reasonable number of lexical substitutes $k$ for a word in a sentence should not exceed 5 or 10. It is often difficult to find many synonyms for a target word in a given context. However, \citep{amrami2019towards} experimentally found optimal $k=200$ for BERT LSDP on WSI benchmarks. We confirm their findings on the train subset of RUSSE'2018 \citeplanguageresource{Russe} for our SDP method, the WSI performance for SDP reaches a plateau only at $k\approx100$ (see Appendix \ref{app:topk}). In the case of +embs, however, we found that $k=20$ is optimal with performance significantly declining for larger values of $k$. As a result, we chose $k=150$ for SDP and $k=20$ for +embs. We further discuss possible reasons for this distinction in section 5. 

The next step in the substitution-based methods is to build a numeric vector for each instance based on the generated substitutes. We lemmatize the substitutes with the Stanza lemmatiztion package\footnote{https://stanfordnlp.github.io/stanza/lemma.html}. It supports 78 of 100 XLM-R languages out of the box. We experimentally compared Stanza to other multilingual and monolingual lemmatization packages and observed no significant impact on WSI results\footnote{We experimented with Spacy (https://spacy.io/api/lemmatizer), WordNet lemmatizer (nltk.stem.wordnet) and PyMorphy2 \citep{pymorphy}. The differences in WSI metrics are within 5\% and are not consistent between datasets and languages.}. We then build a TF-IDF vector from the lemmatized substitutes for each instance.

Finally, TF-IDF vectors get clustered. This step can be considered rather language-independent. While plenty of clustering methods exist, we choose simple agglomerative clustering with cosine distance and average linkage, following previous work on WSI \citep{amrami2019towards, arefyev2020always}. For each target word in the dataset, we perform clustering with the number of clusters ranging from 2 to 9 and select the clustering with the highest Calinski-Harabasz score \citep{calinski1974dendrite}, an unsupervised score of clustering quality. In the soft clustering benchmarks we use a trivial hard-to-soft clustering conversion, where the hard cluster is assigned a probability of 1. While more complicated clustering approaches may increase the overall WSI performance, our primary focus is on the substitute generation, as we consider it to be the most language-dependent step in the lexical-substitution WSI pipeline.

\section{WSI Evaluation}

\subsection{Datasets}\label{41}

\begin{table*}
\footnotesize
\centering

\begin{tabular}{lrrrrr}
 &  &  & \textit{Instances} & \textit{Senses} & \textit{Avg instance} \\
\textit{Dataset} & \textit{Instances} & \textit{words} &  \textit{per word} & \textit{per word} & \textit{length (words)}  \\
\hline
SE10 & 8915 & 100 & 89.2 & 3.9 & 64.9 \\
SE13 & 4664 & 48 & 97.2 & - & 30.3 \\
\hline
bts-rnc-ru-test-private & 4335 & 34 & 127.5 & 3.0 & 25.0 \\
\hline
DeWUG-de-Sense & 826 & 24 & 34.4 & 2.8 & 33.4 \\
\hline
XL-WSD-bg & 3259 & 380.0 & 8.6 & 2.4 & 3.3 \\
XL-WSD-de & 117 & 14.0 & 8.4 & 2.1 & 28.3 \\
XL-WSD-en & 3126 & 396.0 & 7.9 & 2.8 & 26.3 \\
XL-WSD-es & 213 & 36 & 5.9 & 2.2 & 31.4 \\
XL-WSD-fr & 181 & 33 & 5.5 & 2.2 & 34.9 \\
XL-WSD-gl & 972 & 184 & 5.3 & 2.8 & 4.9 \\
XL-WSD-it & 319 & 54 & 5.9 & 2.3 & 31.7 \\
XL-WSD-ko & 865 & 164 & 5.3 & 2.5 & 6.0* \\
XL-WSD-sl & 1711 & 70 & 24.4 & 3.2 & 22.2 \\
XL-WSD-zh & 6774 & 696 & 9.7 & 3.5 & 35.0* \\
\hline
\end{tabular}

\caption{Basic statistics for the used datasets. ISO 639 is used to encode language names.
\\ \footnotesize{* Number of hieroglyphs}}
\label{tab:dataset stats}

\end{table*}

We use the existing WSI datasets as well as datasets created for other lexical semantic tasks for our multilingual WSI experiments. 

The most popular WSI benchmarks for English are \textbf{SE10} (SemEval-2010 task 1, \citealplanguageresource{manandhar-etal-2010-semeval}) and \textbf{SE13} (SemEval-2013 task 13, \citealplanguageresource{jurgens-klapaftis-2013-semeval}). They were annotated manually. SE10 provides gold hard clustering of instances, while SE13 provides gold soft clustering. 

RUSSE'2018 \citeplanguageresource{Russe} is a shared task that presented three manually annotated WSI datasets in Russian: wiki-wiki, bts-rnc and active-dict. We employ the private test subset of \textbf{bts-rnc-ru} in our experiments\footnote{Wiki-wiki dataset contains easy examples and has been solved with a perfect score in the competition and active-dict has a significantly lower instances per target word ratio.}. Bts-rnc-ru also has a predefined train subset, which we use for hyperparameter selection and ablation study.

We use the largest existing Word Sense Disambiguation dataset \textbf{XL-WSD} \citeplanguageresource{XL-WSD} for WSI to extend the number of languages for evaluation. The test subset of XL-WSD provides sense annotations for over 50 thousand instances in 18 languages. We replace the provided sense definitions with cluster labels. We only apply minimal filtering to the test subset of XL-WSD, keeping all the target words with at least two senses and at least three instances. We include only 10 out of 18 languages of XL-WSD because datasets in the other eight languages frequently contain only one instance per sense. We also use train subsets of XL-WSD in English and Slovenian with a stricter filter (at least eight instances per word) as our dev sets. 

We employed the dataset in German \textbf{DWUG de Sense} (\citealplanguageresource{schlechtweg2021dwug}, \citealp{dewugphd}) which, unlike XL-WSD, is human-annotated. Originally, DWUG de Sense is designed to study the change in word usage in particular senses over time. The annotators were given a task to map target words to one of the predefined sense definitions in instances from "old" (XIX century) and "new" (XX century) corpora. We concatenated the "old" and the "new" parts of the dataset with no further processing\footnote{We only removed three instances that are longer than 512 tokens.}. Like in XL-WSD, we replaced sense definitions with cluster labels.

To summarize, we use SE10, SE13, the test subset of bts-rnc-ru, the XL-WSD in 10 languages, and DeWUG de Sense for WSI evaluation. Basic statistics for the used datasets are provided in table \ref{tab:dataset stats}. Notably, the converted datasets (DWUG de Sense and XL-WSD) have a much lower instances per word ratio in comparison to SE10, SE13, and bts-rnc-ru. It makes the task of WSI more challenging, as in many cases clustering must be performed in a very sparse vector space. Also, instances for Bulgarian and Galician in XL-WSD are short. They do not contain full sentences but rather short phrases.

\subsection{Results}

We include a total of 6 configurations in our experiments. We employ four multi-token substitute generators (concat, WCM, WCM-en, and WCM-ru) and combine them with two target injection methods (SDP and +embs). As mentioned above, WCM-en and WCM-ru always yield substitutes in English and Russian, respectively, for context in any language. We exclude WCM-en +embs and WCM-ru +embs from experiments. These configurations require aligned cross-lingual embeddings, which we consider a source of additional error.

\subsubsection{Comparison to Monolingual Methods}
 We compare our methods to existing monolingual ones on established WSI benchmarks: bts-rnc-ru, SE10, and SE13. The results are provided in Table \ref{tab:test eng}. 
 
 Our SDP-based methods perform on par with SOTA English monolingual approaches in hard clustering on SemEval 2010 and do not fall far behind on SemEval 2013 with trivial hard-to-soft clustering conversion. For +embs the results are noticeably worse. We discuss possible explanations for this fact in Section 5. It is worth mentioning that BERT LSDP \citep{amrami2019towards}, PolyLM Large \citep{ansell2021polylm}, XLNET +embs \citep{arefyev2020always}, and our methods have roughly the same number of parameters (350 million) in the underlying neural language models.
 
  In bts-rnc-ru, our systems fall behind only of \citep{xlmr-russe} that uses a fixed number of clusters for all words. This number is fitted on the train set of bts-rnc, and is unlikely to transfer to other languages or datasets. 

Notably, WCM-en is the second best-performing configuration on the Russian benchmark bts-rnc-ru while producing only English substitutes. We confirm that substitutes are indeed in English while the dataset is in Russian by building a simple character-based classifier. Russian and English have different alphabets, therefore we consider a substitute to be in English if all its characters are Latin. Over 99\% of the generated substitutes get classified as English. We also performed a visual inspection, confirming this result. WCM-ru is close to WCM-en on English benchmarks with over 99\% of substitutes in Russian. 

\begin{table}
\footnotesize
\centering
\begin{tabular}{|l|ll|l|}
\hline
\textit{System} & \textit{F-S} & \textit{V-M} & \textit{AVG} \\
\hline
\textbf{Concat SDP} & 67.7 & 41.7 & 53.1 \\
\textbf{Concat +embs} & 62.5 & 32.0 & 44.7 \\
\textbf{WCM SDP} & 65.1 & 39.4 & 50.7 \\
\textbf{WCM +embs} & 64.1 & 26.9 & 41.5 \\
\textbf{WCM-en SDP} & 66.3 & 39.0 & 50.8 \\
\textbf{WCM-ru SDP} & 65.5 & 37.2 & 49.4 \\
BERT LSDP & \textbf{71.3} \tiny{$\pm$0.1} & 40.4 \tiny{$\pm$1.8} & 53.6 \tiny{$\pm$1.2} \\
XLNET +embs & & & 54.2 \\
PolyLM Large* & 67.5 & \textbf{43.6} & \textbf{54.3} \\
AutoSense &  61.7 & 9.8 & 24.5 \\
\hline
\end{tabular}

\begin{tabular}{|l|ll|l|}
\hline
\textit{System} & \textit{FBC} & \textit{FNMI} & \textit{AVG} \\
\hline
\textbf{Concat SDP} & 65.1 & 19.6 & 35.7 \\
\textbf{Concat +embs} & 59.6 & 14.0 & 28.8 \\
\textbf{WCM SDP} & 62.5 & 16.4 & 32.0 \\
\textbf{WCM +embs} & 61.7 & 13.7 & 29.1 \\
\textbf{WCM-en SDP} & 64.0 & 18.8 & 34.6 \\
\textbf{WCM-ru SDP} & 62.2 & 17.4 & 32.9 \\
BERT LSDP & 64.0 \tiny{$\pm$0.5}  & 21.4 \tiny{$\pm$0.5} & 37.0 \tiny{$\pm$0.5} \\
XLNET +embs & & & 37.3 \\
PolyLM Large* & \textbf{66.7} & \textbf{23.7}	& \textbf{39.7} \\
ELMo LSDP & 57.5  & 11.3 & 25.4 \\
AutoSense & 61.7  & 7.96 & 22.2 \\
\hline
\end{tabular}

\begin{tabular}{|l|l|}
\hline
\textit{System} & \textit{ARI} \\
\hline
\textbf{Concat SDP} & 51.6 \\
\textbf{Concat +embs} & 49.1 \\
\textbf{WCM SDP} & 48.9 \\
\textbf{WCM +embs} & 44.2 \\
\textbf{WCM-en SDP} & 50.4 \\
\textbf{WCM-ru SDP} & 50.2 \\
\citet{xlmr-russe} & \textbf{57.3} \\
\citet{arefyev2019combining} & 45.1 \\
RUSSE'2018 competition best [1] & 33.8 \\
RUSSE'2018 competition baseline [1] & 21.3 \\
\hline
\end{tabular}

\caption{ Evaluation results in English on SE10 (top), SE13 (middle) and in Russian on bts-rnc-ru-test-private (bottom).
\\ \footnotesize{* PolyLM Large was not presented in the original paper \citep{ansell2021polylm}, the results are from the authors' official repository.}}
\label{tab:test eng}
\end{table}

\subsubsection{Multilingual WSI evaluation}
 We also evaluate the proposed methods on our custom WSI benchmarks XL-WSD and DWUG de Sense. We employ a trivial adaptation of the substitute generator from the original BERT LSDP approach as a baseline. In the baseline, XLM-R replaces BERT as an underlying model, but multitoken substitutes are not allowed. Like in the original method, the target injection relies on both the unmasked instance and the non-symmetric dynamic pattern. However, we replace the original clustering approach with ours to allow for a better comparison of the substitute generators.
 
\begin{table*}
\footnotesize
\centering

\begin{tabular}{|l|p{0.5cm}p{0.5cm}p{0.7cm}|p{0.9cm}|p{0.4cm}p{0.4cm}p{0.4cm}p{0.4cm}p{0.4cm}p{0.4cm}p{0.4cm}p{0.4cm}p{0.4cm}p{0.4cm}|}
\hline
& SE10 & SE13 & bts-r & DWUG & \multicolumn{10}{c|}{XL-WSD} \\
 & en & en & ru & de & bg & de & en & es & fr & gl & it & ko & sl & zh \\
\hline
 Baseline & 24.2 & 23.5 & 35.4 & 40.4 & 5.0 & 1.1 & 18.4 & 20.5 & 20.6 & 7.6 & 20.7 & 11.6 & 3.9 & 20.4 \\
\hline
 Concat SDP & \underline{\textbf{38.1}} & \underline{\textbf{30.4}} & 47.7 & \textbf{41.9} & \textbf{11.9} & 18.0 & 32.9 & \textbf{34.4} & \textbf{27.8} & \underline{\textbf{11.6}} & \textbf{23.5} & \textbf{24.9} & \underline{\textbf{18.5}} & \underline{\textbf{30.3}} \\
 WCM SDP & 34.9 & 25.4 & \textbf{48.9} & 39.5 & 10.0 & \textbf{23.4} & \underline{\textbf{33.0}} & 29.0 & 27.3 & 10.8 & 22.0 & 17.9 & 13.5 & 26.8 \\
\hline
 WCM-en SDP & 35.8 & 28.8 & \underline{\textbf{50.4}} & 41.2 & \textbf{\underline{18.1}} & 22.7 & 31.9 & 37.0 & 24.7 & 9.3 & \textbf{29.9} & \textbf{\underline{25.6}} & \textbf{17.8} & \textbf{27.8} \\
 WCM-en SDP-en & 35.8 & 28.8 & 47.5& \textbf{\underline{50.8}} & 16.8 & \textbf{\underline{25.1}} & 31.9 & \textbf{38.0} & \textbf{33.1} & \textbf{10.6} & 27.2 & 22.5 & 15.7 & 26.8 \\
\hline
 WCM-ru SDP & \textbf{33.3} & \textbf{28.1} & 50.2 & 46.3 & \textbf{14.8} & \textbf{20.5} & \textbf{31.4} & \textbf{\underline{39.3}} & \textbf{\underline{33.7}} & \textbf{10.5} & \textbf{\underline{30.6}} & \textbf{\underline{25.6}} & \textbf{14.3} & \textbf{27.3} \\
 WCM-ru SDP-ru  & 28.4 & 23.1 & 50.2 & \textbf{46.4} & 13.5 & 19.7 & 30.0 & 33.5 & 21.2 & 10.1 & 24.4 & 23.4 & 12.3 & 25.2 \\
\hline
\end{tabular}

\caption{WSI evaluation results. The metric is Adjusted Rand Index (ARI, \citealp{Hubert1985}). }
\label{tab:xlwsd eval}
\end{table*}

The results on all the test datasets we use for WSI are presented in Table \ref{tab:xlwsd eval}. All of our methods surpass the baseline on each dataset, proving the effectiveness of our approach in a multilingual setup. No configuration is an overall winner.

We include two more configurations in this experiment: \textbf{WCM-en SDP-en} and \textbf{WCM-ru SDP-ru}. These configurations are identical to WCM-en SDP and WCM-ru SDP, respectively, but do not involve the translation of the dynamic patterns. WCM-en SDP-en uses the pattern in English for all languages, and WCM-ru SDP-ru always uses the pattern in Russian. Due to the properties of WCM-en and WCM-ru they also do not require the use of a mulitlingual lemmatizer. We can see from Table \ref{tab:xlwsd eval} that \textbf{WCM-en SDP-en shows strong WSI results and requires no adaptation to new languages.}. We thus consider WCM-en SDP-en to be a truly multilingual WSI system.

\section{Semantic Relations between Substitutes and Target Words}

\begin{figure}
\includegraphics[width=\linewidth]{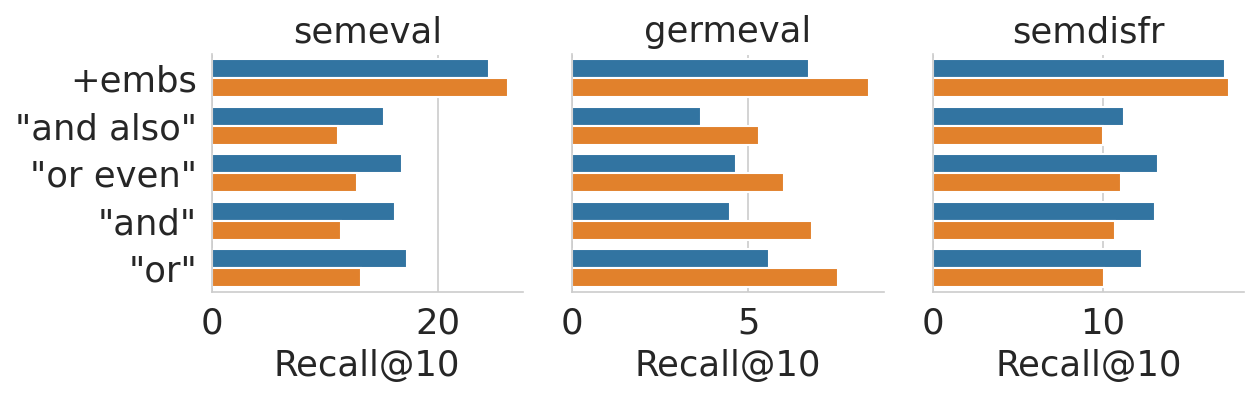}
\caption{Evaluation of substitute generators on the lexical substitution datasets . Blue is the concat substitute generator and orange is the WCM. }
\label{fig:lexsub_plot}
\end{figure}

To get more insights on substitute generation, we also perform a direct evaluation of our substitute generators on existing lexical substitution datasets: SemEval2007 task 10 \citeplanguageresource{Se07} in English, GermEval 2015 \citeplanguageresource{GermEval}  in German and SemDisFr 2014 \citeplanguageresource{SemDis}  in French. The results are presented in Figure \ref{fig:lexsub_plot}. The performance of target injection methods on these datasets is opposite to WSI (in line with the results from \citealp{arefyev2020always}): +embs performs better than SDP "or even" or other dynamic patterns.

To find an explanation for the divergence of results on lexical substitution and WSI, we classify relations between target words and substitutes based on WordNet \citeplanguageresource{wordnet} relations for different target injection methods. WordNet provides "synonym" relations of the synsets as well as hyponymy relations. We classify the relations between the target word and the substitute into the following classes:

    \textit{Direct hyponym}, \textit{direct hypernym}, or \textit{synonym} if the two words have synsets in WordNet with corresponding relations. For examplem "onion" is a direct hyponym of "vegetable", "vegetable" is a direct hypernym of "onion", and "middle" is a synonym of "center".
    
    \textit{Transitive hyponym} or \textit{transitive hypernym} if there is a path of only hyponym or hypernym relations from the target word to the substitute. "Onion" is a transitive hyponym of "food" (onion, vegetable, food), and "food" is a transitive hypernym of "onion".
    
    \textit{Co-hyponym} if two words share a direct hypernym, \textit{co-hyponym-3} if they share a hypernym of a hypernym. "Onion" and "garlic" are co-hyponyms, and "onion" and "apple" are co-hyponyms-3.
    
\begin{figure}
\includegraphics[width=\linewidth]{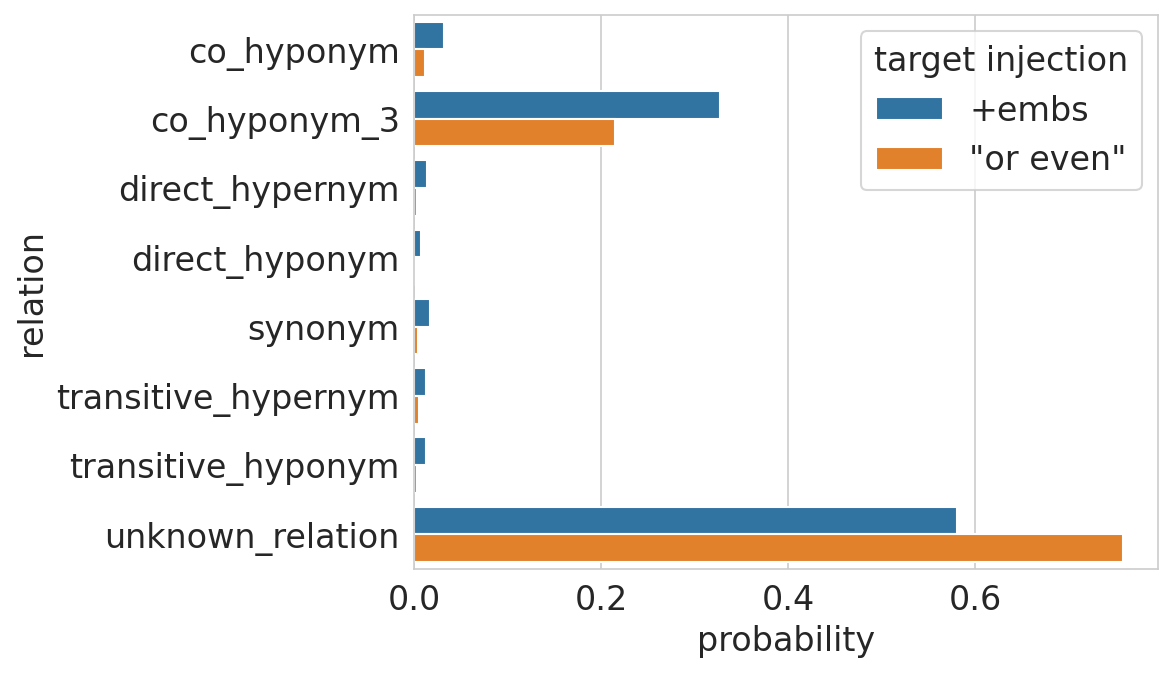}
\caption{Distribution of Wordnet hyponymy relations for different target injection methods. Only top-20 substitutes are used for each instance. }
\label{fig:wordnet_rel}
\end{figure}

Figure \ref{fig:wordnet_rel} represents the distribution of relation classes for "or even" and +embs target injection methods on all instances from SE10 where target words are nouns. SDP "or even" yields substitutes with "unknown relation" more frequently. However, SDP substitutes still allow for better clustering according to WSI evaluation. 

We review the most \textit{discriminative} substitutes for target words in SE10 as a qualitative analysis of the generated substitutes. A substitute is discriminative if it appears frequently in instances for a given sense and rarely in other senses. More formally, the substitutes for each sense get ranked by dividing the number of occurrences of the substitute in this sense by the number of occurrences of that substitute in other senses, according to gold sense clustering. 

\begin{table*}
\centering
\footnotesize

\begin{tabular}{|l|l|l|l|l|l|l|}
\hline
\textit{Target word} & \textit{Context Example} & \textit{Concat +embs} & \textit{Concat SDP} & \textit{WCM-ru SDP}  \\

\hline
\hline

\multirow[c]{3}{*}{\shortstack[l]{body \\ (33 instances)}}
& \multirow[c]{3}{*}{\shortstack[l]{... the highest \\ legislative \textbf{body} \\ of China ...}} 
  & institution & organization & \foreignlanguage{russian}{чиновник}(official) \\
 & & govern & committee &  \foreignlanguage{russian}{учреждение}(establishment) \\
 & & organization & office &  \foreignlanguage{russian}{государство}(state) \\
\hline
\multirow[c]{2}{*}{\shortstack[l]{body \\ (70 instances)}}
& \multirow[c]{2}{*}{\shortstack[l]{... the \textbf{body's} \\ immune response...}} 
   & shape, bodily, & personality, muscle, &  \foreignlanguage{russian}{мышца}(muscle), \foreignlanguage{russian}{взгляд} \\
 & & muscle & health &  (sight), \foreignlanguage{russian}{красота}(beauty) \\
\hline
\multirow[c]{2}{*}{\shortstack[l]{body \\ (78 instances)}} 
& \multirow[c]{2}{*}{\shortstack[l]{ ... his \textbf{body} bore \\ several wounds...}}
   & dead, victim, & victim, debris, &  \foreignlanguage{russian}{имущество}(property), \foreignlanguage{russian}{флаг} \\
 & & remain & grave & (flag), \foreignlanguage{russian}{контейнер}(conatiner) \\
 
\hline
\hline

\multirow[c]{2}{*}{\shortstack[l]{cell \\ (11 instances)}} 
& \multirow[c]{2}{*}{\shortstack[l]{... to its prison \\ \textbf{cell} shortage ...}}
  & jail, room, & apartment, & \foreignlanguage{russian}{коридор}(corridor), \foreignlanguage{russian}{гараж} \\
 & & house & cellar, detention &  (garage), \foreignlanguage{russian}{кабинет}(cabinet) \\
\hline
\multirow[c]{3}{*}{\shortstack[l]{cell \\ (58 instances)}} 
& \multirow[c]{2}{*}{\shortstack[l]{...  of white blood \\ \textbf{cells} involved in the \\ body's immune ... }}
   & tissue  & blood &  \foreignlanguage{russian}{вирус}(virus), \\
 & & tumor & virus &  \foreignlanguage{russian}{организм}(organism) \\
 & & stem & animal &  \foreignlanguage{russian}{новообразование}(tumor) \\
 
\hline
\hline

\multirow[c]{2}{*}{\shortstack[l]{display \\ (11 instances)}}
& \multirow[c]{2}{*}{\shortstack[l]{... on a TV \\ \textbf{display} ...}}
   & lcd, monitor, & LCD, digital,  &  \foreignlanguage{russian}{аналог}(analogue), \foreignlanguage{russian}{дисплей} \\
 & & system & component & (display) \foreignlanguage{russian}{монитор} (monitor) \\
\hline
\multirow[c]{2}{*}{\shortstack[l]{display \\ (19 instances)}}
& \multirow[c]{2}{*}{\shortstack[l]{... diamonds \\ on \textbf{display}...}}
    & sale, showcase, & sell, presentation,  &  \foreignlanguage{russian}{продажа} (sale), \foreignlanguage{russian}{купить}  \\
 &  & view & public & (to buy), \foreignlanguage{russian}{продать} (to sell) \\
\hline
\end{tabular}

\caption{ Top-3 discriminative substitutes for Concat +embs, Concat SDP, and WCM-ru SDP for three nouns from the SE10 dataset. }
\label{tab:disc substs}

\end{table*}

The top 3 discriminative substitutes for several senses of three target words from SE10 are presented in Table \ref{tab:disc substs}. Firstly, we confirm that WCM-ru yields cross-lingual substitutes for instances in the WSI datasets, and these substitutes fit the context quite well. Secondly, we find examples of substitutes that seem useful for WSI but which are not lexical substitutes in a traditional sense. E.g., "\textit{blood}" and "\textit{animal}" indeed help distinguish a "blood \textit{cell}" from a "prison \textit{cell}" but it is hard to describe the exact relation between the target word "cell" and these substitutes.

In general, we conclude that the superior performance of SDPs on WSI can be explained by the presence of words that are not lexical substitutes by conventional definition (they cannot \textit{replace} the target word in the sentence) but tend to co-occur with the target word. It may also explain the optimal WSI $k$ in +embs being lower than in SDP. While +embs is better at producing closely related substitutes, SDP substitutes are more diverse. Words usually do not have many closely related substitutes and +embs quickly "runs out" of adequate substitutes. 

\section{Conclusion}

By harnessing the capabilities of the multilingual masked language model XLM-R and carefully adapting existing monolingual WSI methods, we were able to build a multilingual WSI system that does not require any adaptation to be used in a new language. We performed extensive experiments across several languages and datasets to prove that our approach is robust and performs on par with the state-of-the-art monolingual methods. We also discovered that by simply finetuning XLM-R with the WCM objective in a single language, we obtain a model capable of producing cross-lingual lexical substitutes, and these substitutes allow for adequate WSI performance.

\section{Limitations}
In terms of languages, our system has only one minimal requirement. The language of an instance must be one of 100 languages supported by XLM-R. However, we have only tested our systems on limited academic benchmarks with sense annotations. These benchmarks may not reflect all the nuances of word usage in real corpora. 

Our methods tackle WSI as a hard clustering task, which suggests distinct borders between the senses of an ambiguous word. The existence of such borders is hardly the case for many words in the natural languages \citelanguageresource{jurgens-klapaftis-2013-semeval}. More than that, our WSI systems always consider the target word to be ambiguous, i.e., having at least two senses. We do not tackle the problem of ambiguity detection in our work. Overall, we believe that deeper research into clustering methods for WSI is required to use current WSI systems "in the wild".

\section{Acknowledgements}
Nikolay Arefyev has received funding from the European Union’s Horizon Europe research and innovation program under Grant agreement No 101070350 (HPLT)

\nocite{*}
\section{Bibliographical References}

\bibliographystyle{lrec-coling2024-natbib}
\bibliography{lrec-coling2024-example}

\section{Language Resource References}
\label{lr:ref}
\bibliographystylelanguageresource{lrec-coling2024-natbib}
\bibliographylanguageresource{languageresource}

\appendix

\section{Comparison of multi-token substitute generators and target injection methods on the dev sets}\label{app:hyper_sel}

\begin{figure*}[t]
\includegraphics[width=\linewidth]{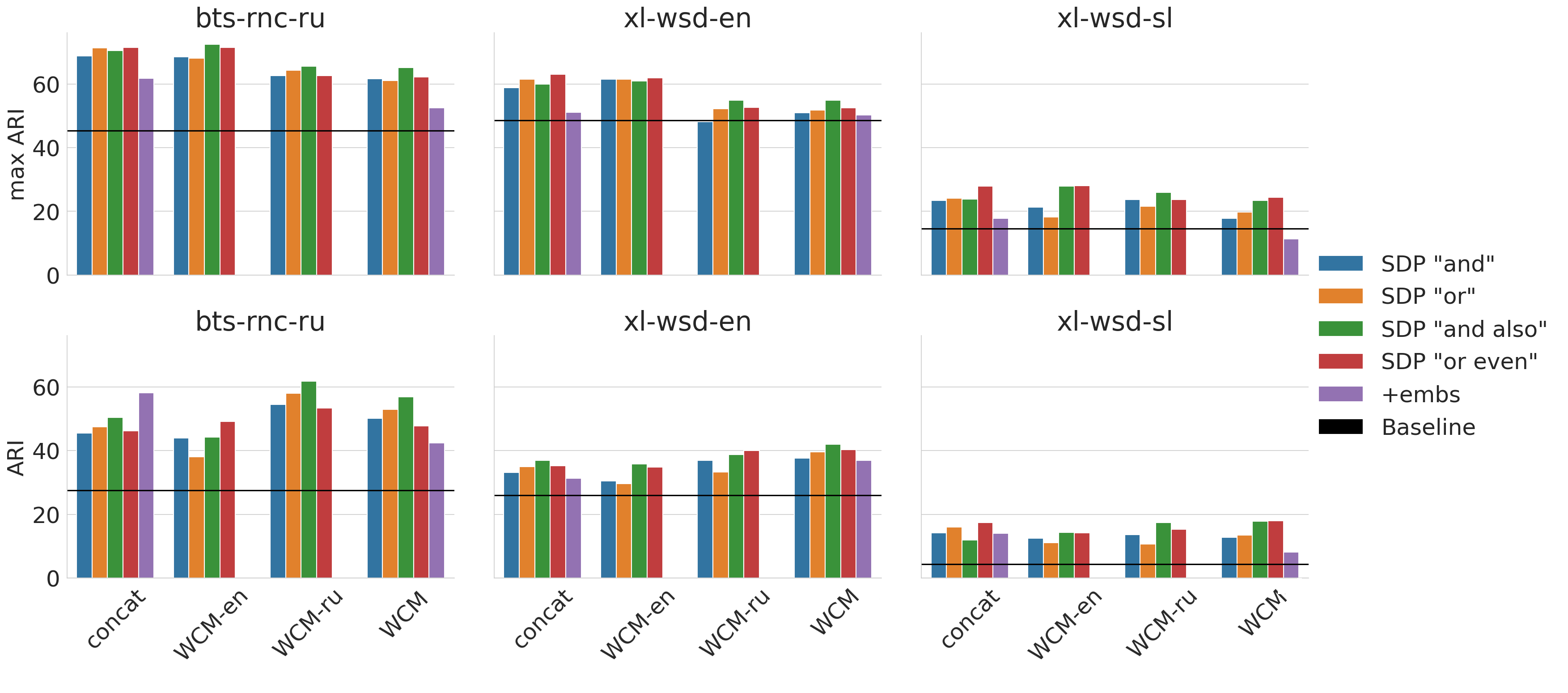}
\caption{Evaluation of different configurations of our system on the WSI dev sets. }
\label{fig:ARI_plot}
\end{figure*}

We evaluated 18 system configurations on the dev sets (russe-bts-rnc-train, XL-WSD-en-train and XL-WSD-sl-train) with different target injection methods (4 dynamic patterns and +embs) and multi-token substitute generators (WCM-multi, WCM-en, WCM-ru, and concat). The used dynamic patterns are presented in Table \ref{tab:sdp}. Baseline method is described in Section 4.2.

\begin{table}[b]
\footnotesize

\begin{center}

\begin{tabular}{||l|l||}

\hline
Name & Symmetric Patterns \\
\hline
"and" & <T> and <mask> \\ & <mask> and <T>\\ 
\hline
"or" & <T> or <mask> \\ & <mask> or <T> \\ 
\hline
"and also" & <T> (and also <mask>) \\ & <mask> (and also <T>) \\
\hline
"or even" & <T> (or even <mask>) \\ & <mask> (or even <T>) \\
\hline
\end{tabular}

\caption{The symmetric dynamic patterns used in our experiments. }
\label{tab:sdp}
\end{center}
\end{table}

Our primary metric, \textbf{ARI} (Adjusted Rand Index, \citealp{Hubert1985}), is a clusterization metric and may vary significantly with changes in clustering hyperparameters (e.g., in the number of clusters). We also measure \textbf{maxARI} to evaluate the quality of substitute vectors themselves. MaxARI is the upper bound for ARI achievable given a vectorizer, a clusterer, and a grid on clustering hyperparameters. It is calculated with the clustering hyperparameters for each target word selected individually to maximize ARI. We observe maxARI to be more stable to variations in substitute generator parameters compared to regular ARI.

The results of the evaluation are presented in Figure \ref{fig:ARI_plot}. The first observation is that all the configurations surpass the baseline in ARI on all the dev datasets. The same can be said for maxARI in 16 out of 18 configurations. In general, SDPs perform better than +embs. We can see that more complicated patterns "and also" and "or even" perform slightly better than simpler patterns "and" and "or".

\section{Selection of number of substitutes}\label{app:topk}

\begin{figure}
\includegraphics[width=\linewidth]{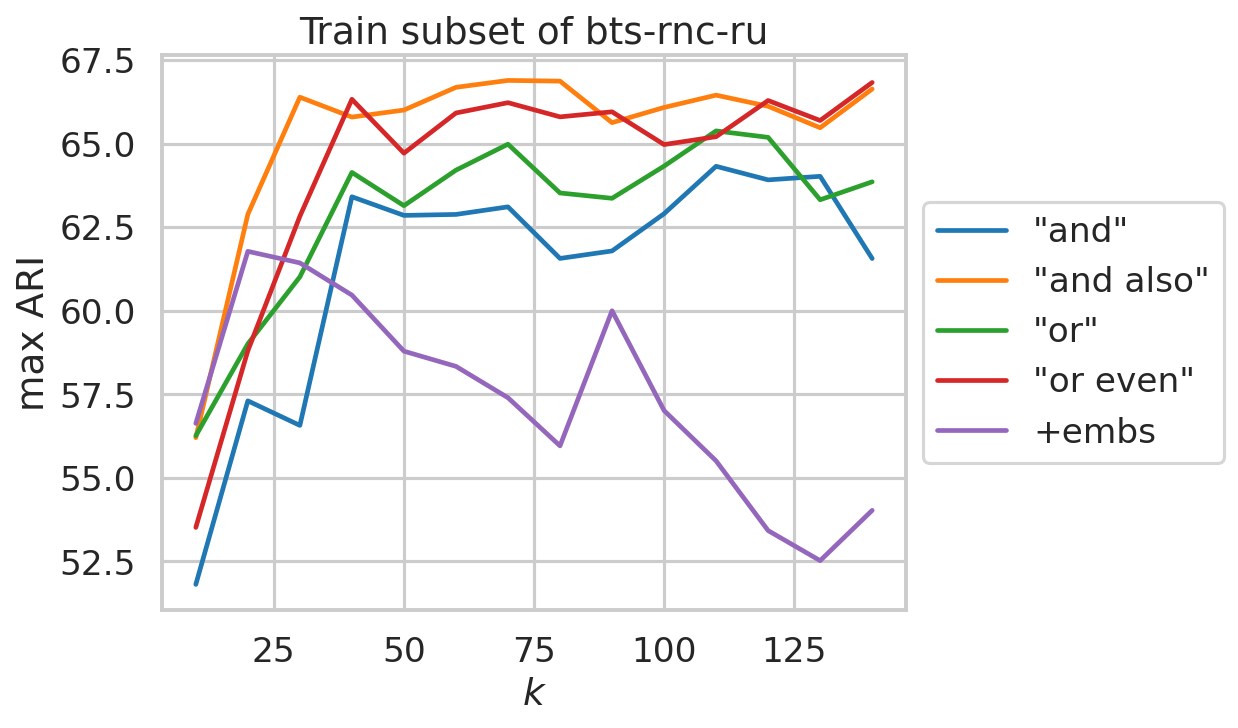}
\caption{Relation between $k$ and maxARI on the train subset of bts-rnc-ru for different target injection methods. Concat is used for multi-token substitute generation. }
\label{fig:topk}
\end{figure}

We optimized number of substitutes ($k$) in our methods based on maxARI on the train subset of bts-rnc-ru for each target injection method. The results are presented in Figure \ref{fig:topk}. The WSI performance for SDP reaches a plateau only at $k\approx100$. In the case of +embs, however, we found that k = 20 is optimal. The performance significantly declines for larger values of $k$.

\section{The importance of multi-token substitutes for WSI}\label{app:num_masks}

\begin{figure}
\includegraphics[width=\linewidth]{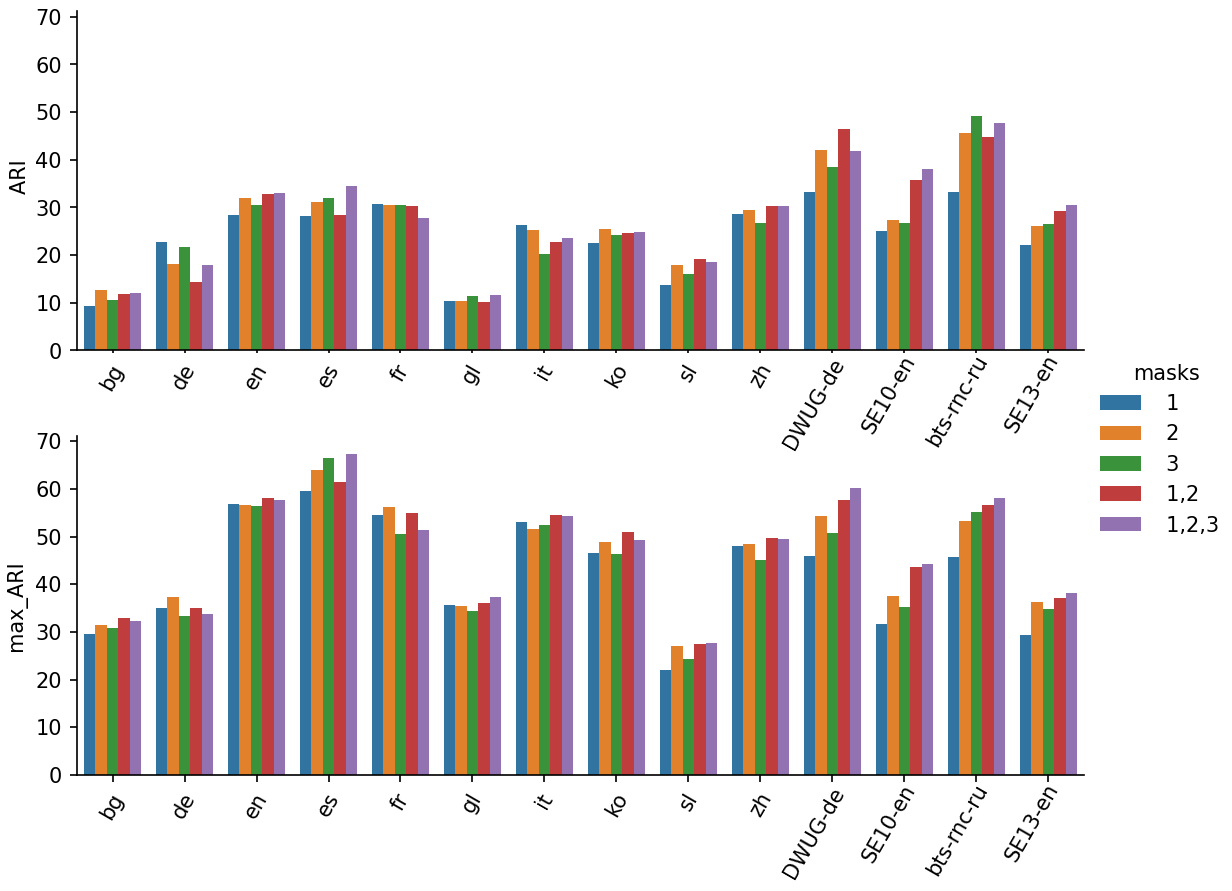}
\caption{The relation between WSI metrics (ARI and maxARI) and the number of masks inserted into the SDP.}
\label{fig:num_masks}
\end{figure}

Figure \ref{fig:num_masks} represents ARI and maxARI with different number of masks inserted into an SDP. We see a notable benefit in using varaible length substitutes (1,2,3 masks in Figure \ref{fig:num_masks} is the concat approach) in SE10, SE13, bts-rnc-ru and DWUG de Sense, but not in XL-WSD.

\section{MaxARI on the test sets}\label{app:maxARI test}

\begin{table*}[!t]
\footnotesize
\centering

\begin{tabular}{|l|p{0.5cm}p{0.5cm}p{0.7cm}|p{0.9cm}|p{0.4cm}p{0.4cm}p{0.4cm}p{0.4cm}p{0.4cm}p{0.4cm}p{0.4cm}p{0.4cm}p{0.4cm}p{0.4cm}|}
\hline
& SE10 & SE13 & bts-r & DWUG & \multicolumn{10}{c|}{XL-WSD} \\
 & en & en & ru & de & bg & de & en & es & fr & gl & it & ko & sl & zh \\
\hline
 Baseline & 24.2 & 23.5 & 35.4 & 40.4 & 5.0 & 1.1 & 18.4 & 20.5 & 20.6 & 7.6 & 20.7 & 11.6 & 3.9 & 20.4 \\
\hline
 Concat SDP & \underline{\textbf{38.1}} & \underline{\textbf{30.4}} & 47.7 & \textbf{41.9} & \textbf{11.9} & 18.0 & 32.9 & \textbf{34.4} & \textbf{27.8} & \underline{\textbf{11.6}} & \textbf{23.5} & \textbf{24.9} & \underline{\textbf{18.5}} & \underline{\textbf{30.3}} \\
 WCM SDP & 34.9 & 25.4 & \textbf{48.9} & 39.5 & 10.0 & \textbf{23.4} & \underline{\textbf{33.0}} & 29.0 & 27.3 & 10.8 & 22.0 & 17.9 & 13.5 & 26.8 \\
\hline
 WCM-en SDP & 35.8 & 28.8 & \underline{\textbf{50.4}} & 41.2 & \textbf{\underline{18.1}} & 22.7 & 31.9 & 37.0 & 24.7 & 9.3 & \textbf{29.9} & \textbf{\underline{25.6}} & \textbf{17.8} & \textbf{27.8} \\
 WCM-en SDP-en & 35.8 & 28.8 & 47.5& \textbf{\underline{50.8}} & 16.8 & \textbf{\underline{25.1}} & 31.9 & \textbf{38.0} & \textbf{33.1} & \textbf{10.6} & 27.2 & 22.5 & 15.7 & 26.8 \\
\hline
 WCM-ru SDP & \textbf{33.3} & \textbf{28.1} & 50.2 & 46.3 & \textbf{14.8} & \textbf{20.5} & \textbf{31.4} & \textbf{\underline{39.3}} & \textbf{\underline{33.7}} & \textbf{10.5} & \textbf{\underline{30.6}} & \textbf{\underline{25.6}} & \textbf{14.3} & \textbf{27.3} \\
 WCM-ru SDP-ru  & 28.4 & 23.1 & 50.2 & \textbf{46.4} & 13.5 & 19.7 & 30.0 & 33.5 & 21.2 & 10.1 & 24.4 & 23.4 & 12.3 & 25.2 \\
\hline
\hline
 Baseline & 34.9 & 31.8 & 46.0 & 51.5 & 21.8 & 26.6 & 47.8 & 50.7 & 50.3 & 31.2 & 49.3 & 38.2 & 13.5 & 41.1 \\
\hline
 Concat SDP & \textbf{44.2} & \underline{\textbf{38.2}} & \textbf{58.0} &\textbf{60.3} & \textbf{32.3} & \textbf{33.8} &\textbf{\underline{57.6}} & \textbf{\underline{67.7}} & 51.5 & \underline{\textbf{37.4}} & \textbf{54.4} & \textbf{49.3} & \textbf{27.7} & \textbf{\underline{49.5}} \\
 WCM SDP & 43.3 & 35.4 & 56.8 & 57.3 & 27.6 & 33.4 & 56.9 & 58.4 & \textbf{53.7} & 33.6 & 54.0 & 41.2 & 23.9 & 45.1 \\
\hline
 WCM-en SDP & \underline{44.6} & 37.9 & \underline{\textbf{60.5}} & 60.7 & \textbf{\underline{38.3}} & 35.5 & 57.3 & \textbf{65.0} & 53.8 & 36.7 & \textbf{\underline{60.1}} & \textbf{48.4} & \textbf{\underline{28.3}} & 46.3 \\
 WCM-en SDP-en & \underline{44.6} & 37.9 & 57.1 & \textbf{\underline{64.8}} & 38.0 & \textbf{\underline{36.9}} & 57.3 & 63.6 & \textbf{59.6} & \textbf{37.3} & 56.5 & 46.7 & 24.8 & \textbf{46.6} \\
\hline
 WCM-ru SDP & \textbf{41.7} & \textbf{36.8} & 58.1 & \textbf{60.6} & \textbf{34.9} & \textbf{36.2} & 55.1 & \textbf{66.8} & \textbf{\underline{64.8}} & \textbf{36.2} & \textbf{53.9} & \textbf{\underline{50.3}} & \textbf{25.0} & \textbf{46.4} \\
 WCM-ru SDP-ru & 40.0 & 33.4 & 58.1 & 58.1 & 32.4 & 36.1 & \textbf{56.1} & 60.1 & 51.5 & 35.1 & 52.2 & 45.5 & 23.0 & 44.1 \\
\hline
\end{tabular}

\caption{Extended WSI evaluation results. The metrics are ARI (top) and maxARI (bottom).}
\label{tab:xlwsd eval ext}
\end{table*}

Table \ref{tab:xlwsd eval ext} extends Table \ref{tab:xlwsd eval} and includes maxARI evaluation for the tests. Overall, the differnces in maxARI between configurations are more subtle, compared to regular ARI. The gap between ARI and maxARI is more noticeable for XL-WSD, than for other benchmarks. It means that Calinski-Harabasz score is worse at determining the optimal number of clusters in XL-WSD compared to other datasets. We suggest that it can be explained by differences in statistics for these datasets, highlighted in Section \ref{41}.

\end{document}